\documentclass[11pt]{article}
\usepackage[margin=1in]{geometry}
\usepackage[T1]{fontenc}
\usepackage{lmodern}
\usepackage{amsmath,amssymb}
\usepackage{booktabs}
\usepackage{graphicx}
\usepackage{xcolor}
\usepackage[hidelinks]{hyperref}
\usepackage{enumitem}
\usepackage{flafter}
\usepackage{tikz}
\usetikzlibrary{arrows.meta,positioning}
\usepackage{pgfplots}
\pgfplotsset{compat=1.17}

\newcommand{\sys}{Glance}

\title{\sys: The Generator Is Its Own Reranker}
\author{Adilet Metinov \qquad Chingiz Arziev\\
Airun Intelligence Labs}
\date{October 2026}

\begin{document}
\maketitle

\begin{abstract}
Retrieval-augmented generation commonly serves separate models for reranking and answering, duplicating model residency and passage processing. We introduce \sys{}, which equips the generator with a lightweight decision adapter and pointer head to score candidate passages from a shared prefill. Short relevance probes branch from the cached context, producing one score per passage without separately prefilling each query--passage pair. A single resident 9B backbone supports generation, general decisions and reranking, with approximately 0.1\,GiB of additional memory per decision skill. After 39 minutes of MS MARCO fine-tuning, \sys{} achieves nDCG@10 of 0.741 on SciFact and 0.345 on NFCorpus, compared with 0.721 and 0.326 for bge-reranker-v2-m3. On 300 HotpotQA questions, its selected context improves answer F1 by 0.045 over cross-encoder-selected context under the same base generator (paired bootstrap 95\% CI $[0.014, 0.076]$). In a scaling study, reranking quality keeps rising up to 200 candidates (43k tokens) scored from one shared prefill, where it exceeds the cross-encoder by 0.058 nDCG@10, and a probe-only activated adapter lets generation continue from the original passage cache. These initial results demonstrate generator-based reranking with one resident backbone; evaluation currently covers two retrieval datasets, one QA dataset and one backbone family. Adapters and code: \url{https://huggingface.co/airun-ai/GlanceRAG-MiMo-9B-Reranker}.
\end{abstract}

\section{Introduction}
RAG systems often assign evidence selection and answer generation to separate models. A cross-encoder reranker encodes each query--passage pair separately, typically in batches~\cite{nogueira2020monot5,chen2024bgem3}; LLM-based rerankers improve quality further~\cite{sun2023rankgpt,pradeep2023rankzephyr,zhang2025qwen3emb} at higher cost~\cite{an2025hyperrag}. Either way, an additional model must be served and the candidate passages are processed before the generator reads the selected context. \emph{The generator is going to read the passages anyway: can it judge all of them using computation shared across passages?}

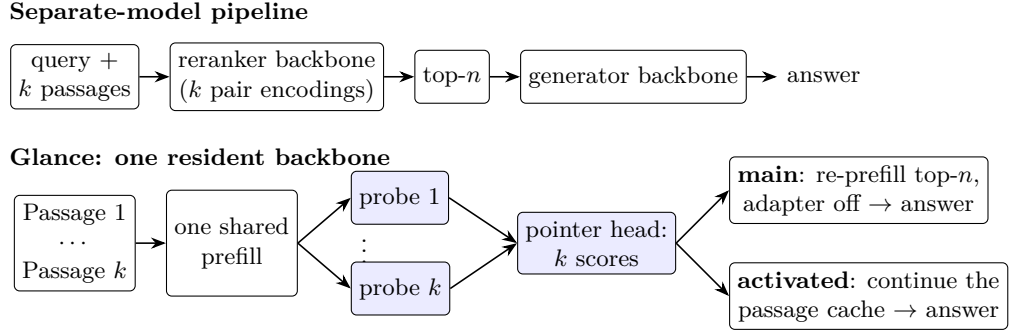
\begin{figure}[t]
\centering
\begin{tikzpicture}[font=\footnotesize, box/.style={draw, rounded corners=2pt, align=center, minimum height=7mm, inner sep=3pt}, ad/.style={box, fill=blue!8}, arr/.style={-{Stealth[length=2mm]}, semithick}, node distance=4mm]
\node[font=\footnotesize\bfseries, anchor=west] (t1) at (0,1.15) {Separate-model pipeline};
\node[box] (q1) at (1.0,0.3) {query +\\$k$ passages};
\node[box, right=of q1] (rr) {reranker backbone\\($k$ pair encodings)};
\node[box, right=of rr] (top1) {top-$n$};
\node[box, right=of top1] (gen1) {generator backbone};
\node[right=of gen1] (ans1) {answer};
\draw[arr] (q1) -- (rr); \draw[arr] (rr) -- (top1); \draw[arr] (top1) -- (gen1); \draw[arr] (gen1) -- (ans1);
\node[font=\footnotesize\bfseries, anchor=west] (t2) at (0,-0.75) {\sys{}: one resident backbone};
\node[box] (q2) at (1.0,-1.9) {Passage 1\\$\cdots$\\Passage $k$};
\node[box, right=of q2, minimum height=14mm] (pre) {one shared\\prefill};
\node[ad, right=7mm of pre, yshift=6mm] (p1) {probe 1};
\node[right=7mm of pre, yshift=0mm] (pd) {$\vdots$};
\node[ad, right=7mm of pre, yshift=-6mm] (pk) {probe $k$};
\node[ad, right=9mm of p1, yshift=-6mm] (head) {pointer head:\\$k$ scores};
\draw[arr] (q2) -- (pre); \draw[arr] (pre.east) -- (p1.west); \draw[arr] (pre.east) -- (pk.west);
\draw[arr] (p1.east) -- (head.west); \draw[arr] (pk.east) -- (head.west);
\node[box, right=7mm of head, yshift=7mm] (main) {\textbf{main}: re-prefill top-$n$,\\adapter off $\rightarrow$ answer};
\node[box, right=7mm of head, yshift=-7mm] (act) {\textbf{activated}: continue the\\passage cache $\rightarrow$ answer};
\draw[arr] (head.east) -- (main.west); \draw[arr] (head.east) -- (act.west);
\end{tikzpicture}
\caption{One resident backbone, shared passage encoding, passage-specific decisions. Shaded parts run with the decision adapter (LoRA + pointer head). In the main pipeline the selected passages are re-prefilled and answered with the adapter off; in the activated variant the passages are prefilled with the base weights, so generation continues the same cache, which still holds all $k$ candidates.}
\label{fig:arch}
\end{figure}

\sys{} equips the generator with a lightweight decision adapter. It prefills the candidate passages once, then evaluates short passage-specific probes that branch from the shared cache, and a 256-dimensional pointer head turns each probe into a relevance score. The same resident backbone then generates the answer (Figure~\ref{fig:arch}). In an activated-adapter variant, passage encoding uses the base weights, so generation can continue from the original cache.

Our contributions:
\begin{enumerate}[leftmargin=*,itemsep=2pt]
\item \textbf{Method.} Passage-specific relevance probes over one shared passage prefill, read by the generator's own backbone through a small adapter and pointer head, including on hybrid linear/full-attention backbones (\S\ref{sec:method}).
\item \textbf{Evidence.} Retrieval results on two BEIR datasets, zero-shot and after short MS MARCO training; downstream QA on HotpotQA under the same generator; and scaling to 200 candidates per shared prefill (\S\ref{sec:exp}).
\item \textbf{Unified serving.} Small decision adapters sharing one backbone, switched per request with outputs identical to each model loaded alone, and an activated variant whose answer continues from the reranking cache (\S\ref{sec:multi}, \S\ref{sec:alora}).
\end{enumerate}
\sys{} removes the additional reranker backbone while delivering competitive retrieval quality and improved downstream answers in the evaluated settings; a dedicated cross-encoder remains faster per query.

\section{Related Work}
\paragraph{Rerankers.} Cross-encoders such as monoT5~\cite{nogueira2020monot5} and BGE rerankers~\cite{chen2024bgem3} score each query--passage pair independently. LLM rerankers score pointwise via yes/no logits~\cite{zhang2025qwen3emb}, or listwise via permutation generation~\cite{sun2023rankgpt,pradeep2023rankzephyr}; FIRST~\cite{reddy2024first} reads the ranking from the logits of the first generated token. In-context reranking~\cite{chen2024icr} uses attention patterns of a frozen LLM over all passages at once. Unlike in-context reranking (attention heuristics, frozen weights) and listwise methods (one permutation or one token for all candidates), \sys{} attaches a separate trained probe per passage over one shared multi-passage context and reads $k$ passage-specific relevance scores; unlike Self-RAG~\cite{asai2023selfrag}, which judges each passage in its own pass, all judgments come from one shared read.

\paragraph{Shared-context and cache reuse.} Parallel Context Windows~\cite{ratner2023pcw} restart positions per window to process segments independently. TurboRAG~\cite{lu2024turborag} and CacheBlend~\cite{yao2024cacheblend} precompute and fuse passage KV caches for generation. HyperRAG~\cite{an2025hyperrag} caches document-side KV offline for decoder rerankers and reports 2--3$\times$ throughput with no loss in accuracy. The Kev runtime~\cite{kev2026} packs many typed questions over one shared state, continuing each question branch from the state's cache, including on hybrid Qwen3.5-style backbones; we build on it and apply it to reranking inside RAG. Activated LoRA~\cite{greenewald2025alora} applies an adapter only after an invocation sequence so that the base model's KV cache can be reused for generation; our main pipeline re-prefills the selected passages (\S\ref{sec:hybrid}), and \S\ref{sec:alora} adopts activation to reuse the cache.

\paragraph{Retrieval-generation coupling.} REPLUG~\cite{shi2023replug} trains a retriever from the LM's own likelihood; Self-RAG~\cite{asai2023selfrag} lets the generator critique retrieved passages through reflection tokens. Long-context models still miss evidence placed in the middle of long inputs~\cite{liu2024lost}, which motivates reranking even when the context window is large.

\section{Method}\label{sec:method}
\subsection{Overview and the two answer pipelines}
\sys{} has two variants, which differ only in where the adapter is active and how the answer is produced (Figure~\ref{fig:arch}).
\begin{itemize}[leftmargin=*,itemsep=2pt]
\item \textbf{\sys{} (main).} The decision adapter is active on all tokens. Prefill the candidates $\rightarrow$ score the probes $\rightarrow$ select the top-$n$ passages $\rightarrow$ re-prefill the selected passages with the adapter off $\rightarrow$ answer.
\item \textbf{\sys{}-Activated.} The adapter is active only on probe tokens. Prefill the candidates with the base weights $\rightarrow$ score the probes with the adapter $\rightarrow$ continue the original cache with the base weights $\rightarrow$ answer. The continued cache retains all $k$ candidates; losing passages are not removed from it.
\end{itemize}
All experiments use the main pipeline except \S\ref{sec:alora}. Table~\ref{tab:config} lists the adapter configurations.

\begin{table}[t]
\centering
\footnotesize
\begin{tabular}{@{}llll@{}}
\toprule
Name & Initialised from & Training & Adapter active on\\
\midrule
zero-shot & -- & 4.4k general decision records, 20 min, no retrieval data & all tokens\\
trained & zero-shot & + 4{,}000 MS MARCO records, 8--20 passages, 39 min & all tokens\\
trained-long & trained & + 450 MS MARCO records, 20--80 passages, 14 min & all tokens\\
control & zero-shot & + 1{,}200 MS MARCO records, 150 steps & all tokens\\
activated & zero-shot & + 1{,}200 MS MARCO records, 150 steps & probe tokens only\\
\bottomrule
\end{tabular}
\caption{Adapter configurations (rank-16 LoRA + 256-d pointer head on the frozen 9B backbone).}
\label{tab:config}
\end{table}

\subsection{Decision mode}
Given a state $s$ and a question with options $o_1,\dots,o_m$, the decision model encodes \texttt{<state> $s$ <q> instruction <opt> $o_1$ </opt> \dots{} <decide>}. The state is the shared context prefix that all questions continue from. A pointer head compares the hidden state at \texttt{<decide>} with the hidden state at each \texttt{</opt>} and returns a softmax over options. Binary yes/no questions (Kev's \texttt{noul} type) use the options \{false, true\}. The head and a rank-16 LoRA~\cite{hu2022lora} adapter are the only trained parameters; the backbone is frozen. No text is generated, so a decision costs one prefill.

\subsection{Multi-probe packing}
For a query $q$ and candidates $p_1,\dots,p_k$ we compare three packings:
\begin{itemize}[leftmargin=*,itemsep=2pt]
\item \textbf{packed\_all}: state $=$ ``[Passage 1] $p_1$ \dots{} [Passage $k$] $p_k$''; branch $i$ is the short question ``Does Passage $i$ contain information that answers or supports the query: $q$?''.
\item \textbf{packed\_q}: state $=q$; branch $i$ contains passage $p_i$ and a relevance question.
\item \textbf{listwise}: state $=q$; one choice question whose options are the $k$ passages.
\end{itemize}
Branches never attend to each other; each branch continues from the end of the state, and its positions continue from there. As shorthand, ``single pass'' means one shared prefill of the passages followed by one batched continuation over all $k$ probes ($\sim$30 tokens each), so no passage is prefilled more than once. The probes give $k$ passage-specific relevance scores; they are not statistically independent, since in packed\_all every probe sees all passages, which lets it compare them.

\subsection{Hybrid backbones}\label{sec:hybrid}
Our 9B backbone interleaves Gated DeltaNet~\cite{yang2025gdn} linear-attention layers with full-attention layers (3:1). An attention mask can isolate branches in full-attention layers, but a linear-attention layer carries a single recurrent state through the sequence. Kev's shared-prefix path therefore runs the shared state once and starts every branch from a copy of each layer's cache at the end of the state: keys and values for attention layers, and the convolution window plus recurrent state for DeltaNet layers (the chunked delta rule's \texttt{initial\_state}). This is exact, and the same path is used in training, so the state is computed once per record rather than once per question. Unlike attention KV, a DeltaNet state summarises the whole prefix in order, so per-passage caches cannot be stitched in arbitrary order; the main pipeline therefore re-prefills the selected passages (about 1.5k tokens), also because its state cache was computed with the adapter on, while \sys{}-Activated continues the full candidate cache.

\subsection{One resident backbone}
The adapter is kept unmerged. Reranking runs with the adapter and pointer head; answer generation runs the same backbone with the adapter disabled, using the top-$n$ passages and the model's chat template. Additional decision skills cost approximately 0.1\,GiB each (LoRA plus head); inference memory also depends on the candidate count and the cache implementation (\S\ref{sec:scale}).

\subsection{Training}\label{sec:train}
We fine-tune from an existing general decision adapter on 4{,}000 MS MARCO~\cite{bajaj2016msmarco} training queries (Tevatron BM25 hard negatives). Each record packs one positive and 7--19 hard negatives, shuffled, in the packed\_all format, with one binary label per probe. One epoch (500 optimizer steps, 55{,}598 labelled probes, 7.7M forward tokens), LoRA rank 16, learning rate $5\cdot10^{-5}$ (head $10^{-4}$), batch 8, bf16, on one NVIDIA B200 in 39 minutes with 21\,GiB peak GPU memory. No BEIR or HotpotQA data is used in training.

\section{Experiments}\label{sec:exp}
\subsection{Setup}
\textbf{Backbone.} MiMo-V2.6-Distill-Qwen-9B (XiaomiMiMo), a hybrid Gated DeltaNet/full-attention model (3:1), frozen in bf16. \textbf{Zero-shot adapter.} ``Zero-shot'' denotes the general decision adapter (rank-16 LoRA + pointer head), trained for 20 minutes on 4.4k public and synthetic decision records (e.g.\ BoolQ, topic and sentiment classification, paraphrase, sentence-level answer containment); the mixture contains no MS MARCO, BEIR or HotpotQA data. \textbf{Retrieval.} BM25 (bm25s) top-$k{=}20$ over each corpus; passages truncated to 220 words. \textbf{Datasets.} BEIR~\cite{thakur2021beir} SciFact~\cite{wadden2020scifact} (300 test queries) and NFCorpus~\cite{boteva2016nfcorpus} (323 test queries), nDCG@10. \textbf{Baselines.} BM25 order; \texttt{bge-reranker-v2-m3} cross-encoder~\cite{chen2024bgem3}; the same decision model scoring each passage in a separate, sequential call (\emph{separate}, passage as state, query as question). \textbf{Hardware.} One RTX PRO 6000 (96 GB) for zero-shot runs; latencies are median per query, bf16, without CUDA graphs or fused kernels, on a GPU shared with other jobs, so latencies may differ under isolated execution and ratios between methods are approximate. Training, the trained-adapter evaluations and \S\ref{sec:scale} and \S\ref{sec:vllm} use one B200, and \S\ref{sec:alora} one H100; latencies are comparable only within a section. BEIR differences are not significance-tested; at $n\approx300$, gaps below about 0.02 nDCG@10 should be read as indicative.

\subsection{Single pass vs.\ \texorpdfstring{$k$}{k} passes (zero-shot)}\label{sec:e1}
\begin{table}[t]
\centering
\begin{tabular}{lcccc}
\toprule
 & \multicolumn{2}{c}{SciFact} & \multicolumn{2}{c}{NFCorpus}\\
Method & nDCG@10 & ms/query & nDCG@10 & ms/query\\
\midrule
BM25 & 0.662 & -- & 0.306 & --\\
bge-reranker-v2-m3 & 0.721 & 31 & 0.326 & 60\\
Decision 9B, separate ($k$ calls) & \textbf{0.729} & 2210 & 0.323 & 2485\\
\sys{} packed\_all (1 pass) & 0.715 & 614 & \textbf{0.341} & 670\\
\sys{} packed\_q (1 pass) & 0.721 & 773 & 0.328 & 847\\
\sys{} listwise (1 pass) & 0.684 & 544 & 0.331 & 573\\
\bottomrule
\end{tabular}
\caption{Reranking BM25 top-20 with the zero-shot adapter (no retrieval training data). One-pass rows run one prefill of the shared state plus one batched continuation over its branches per query; latencies on a shared RTX PRO 6000.}
\label{tab:e1}
\end{table}
Without any retrieval training, every one-pass mode improves on BM25, and on SciFact the shared-passages mode (packed\_all) keeps most of the quality of 20 separate decision calls (0.715 vs.\ 0.729 nDCG@10) 3.6$\times$ faster than sequential pointwise calls to the same 9B model in this shared-GPU experiment, close to the dedicated cross-encoder (0.721). On NFCorpus, packed\_all scores highest in Table~\ref{tab:e1} (0.341, within 0.02 of listwise and packed\_q), above the cross-encoder (0.326) and above the same model's 20 isolated calls (0.323), at 3.7$\times$ lower latency than the latter; one plausible explanation is that each probe can compare a passage with its neighbours, which isolated scoring cannot do, although on SciFact the isolated calls are slightly better. The cross-encoder remains far faster per query (31--60\,ms). In a 20-query pilot on SciFact with less GPU contention, packed\_all scored 0.699 nDCG@10 against 0.697 for 20 separate calls at half the latency (0.60\,s vs.\ 1.20\,s), so the speed-up ratio depends on load; on Kev's fp32-exact evaluation path, which recomputes the state for every question, the same packed request took 16.2\,s, so computing the shared state once is what makes single-pass reranking practical.

\subsection{Training for reranking}\label{sec:e3}
We fine-tune the decision adapter for 39 minutes on MS MARCO in the packed\_all format (\S\ref{sec:train}) and re-evaluate without any BEIR or HotpotQA training data (Table~\ref{tab:e3}). Latencies in this section were measured on a B200.

\begin{table}[t]
\centering
\begin{tabular}{lcccc}
\toprule
 & \multicolumn{2}{c}{SciFact} & \multicolumn{2}{c}{NFCorpus}\\
Method & zero-shot & trained & zero-shot & trained\\
\midrule
BM25 & \multicolumn{2}{c}{0.662} & \multicolumn{2}{c}{0.306}\\
bge-reranker-v2-m3 & \multicolumn{2}{c}{0.721} & \multicolumn{2}{c}{0.326}\\
\sys{} packed\_all (1 pass) & 0.715 & \textbf{0.741} & 0.341 & \textbf{0.345}\\
\sys{} packed\_q (1 pass) & 0.721 & 0.738 & 0.328 & 0.338\\
\sys{} listwise (1 pass) & 0.684 & 0.692 & 0.331 & 0.334\\
\bottomrule
\end{tabular}
\caption{nDCG@10 before and after 39 minutes of MS MARCO training (BM25 top-20). Trained packed\_all latency on a B200: 420\,ms (SciFact), 411\,ms (NFCorpus) per query.}
\label{tab:e3}
\end{table}

The trained-long adapter of \S\ref{sec:scale} (14 more minutes on records with 20--80 passages) scores 0.749 (SciFact) and 0.346 (NFCorpus) in the same setting, so training on longer contexts does not hurt the standard $k{=}20$ case. After training, packed\_all is the best method in Table~\ref{tab:e3} on both datasets (not significance-tested; separate per-passage calls were not re-run with the trained adapter), above the cross-encoder by 0.020 (SciFact) and 0.019 (NFCorpus) nDCG@10, and above the zero-shot model's 20 separate calls on SciFact (0.729).

\subsection{How far does one pass scale?}\label{sec:scale}
We vary the number of candidates read in one pass, $k\in\{10,20,50,100,200\}$, on the first 100 SciFact queries with passages trimmed to 100 words (Table~\ref{tab:scale}). The zero-shot adapter peaks at $k{=}20$ and degrades at $k{=}100$. The MS MARCO-trained adapter, although trained only on 8--20 passages per record, instead \emph{improves} with $k$: at $k{=}100$ (21.6k context tokens in one pass) it reaches 0.682 nDCG@10, against 0.637 for the cross-encoder scoring the same 100 candidates. A further 14 minutes of training on records with 20--80 passages (\emph{trained-long}) raises every evaluated $k$ (20--100) again, to 0.697 at $k{=}100$ (+0.060 over the cross-encoder, 1.05\,s per query on a B200). At $k{=}200$ (43.4k tokens), which no longer fits in one call and runs on Kev's chunked path (one prefill of the state, probe rows in several batches; 15.8\,s per query), it reaches 0.717 against 0.659 for the cross-encoder on the same 100 queries. Training thus prevents the observed degradation through $k{=}100$ in this study, which suggests that it is not an inherent limit of shared-prefill reranking. The cross-encoder scores lower here than in Table~\ref{tab:e1}, presumably because of the 100-query subset and the shorter passages.

\begin{table}[t]
\centering
\begin{tabular}{rrccccrr}
\toprule
$k$ & tokens/pass & zero-shot & trained & trained-long & bge-v2-m3 & ms/query & peak GiB\\
\midrule
10 & 2.2k & 0.651 & 0.653 & -- & 0.626 & 252 & 15.4\\
20 & 4.3k & 0.662 & 0.658 & 0.674 & 0.624 & 165 & 16.7\\
50 & 10.8k & 0.637 & 0.674 & 0.691 & 0.642 & 417 & 25.5\\
100 & 21.6k & 0.584 & 0.682 & 0.697 & 0.637 & 1049 & 56.4\\
200$^\dagger$ & 43.4k & -- & -- & \textbf{0.717} & 0.659 & 15790 & 20.5\\
\bottomrule
\end{tabular}
\caption{Single-pass reranking with $k$ candidates (SciFact, 100 queries, BM25 top-$k$, 100-word passages). \emph{Trained}: MS MARCO adapter of \S\ref{sec:train} (8--20 passages per record). \emph{Trained-long}: warm-started from it and trained 14 more minutes on 450 records with 20--80 passages. Latency and memory: trained-long on a B200 ($k{=}10$: trained adapter, separate run). $^\dagger$Does not fit in one call; run on Kev's chunked path (one prefill of the state, probe rows in several batches).}
\label{tab:scale}
\end{table}

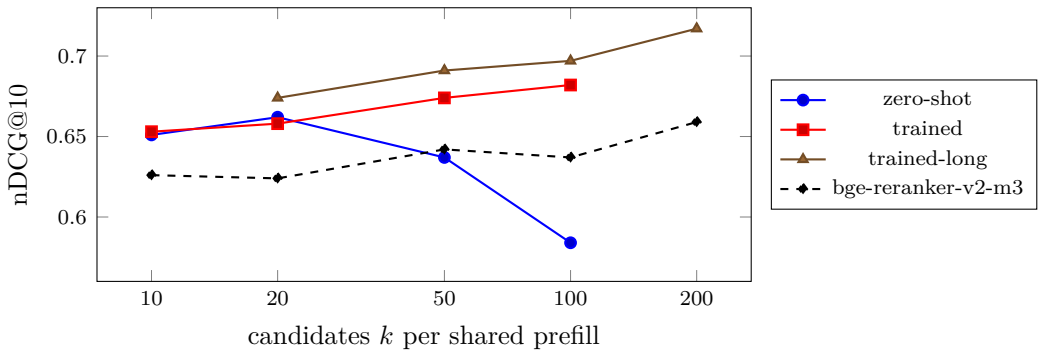
\begin{figure}[t]
\centering
\begin{tikzpicture}
\begin{axis}[width=0.62\linewidth, height=5.2cm, xmode=log, log basis x=10, xtick={10,20,50,100,200}, xticklabels={10,20,50,100,200}, xlabel={candidates $k$ per shared prefill}, ylabel={nDCG@10}, ymin=0.56, ymax=0.73, legend style={font=\scriptsize, at={(1.03,0.5)}, anchor=west}, tick label style={font=\scriptsize}, label style={font=\small}]
\addplot+[mark=*, thick] coordinates {(10,0.651) (20,0.662) (50,0.637) (100,0.584)};
\addplot+[mark=square*, thick] coordinates {(10,0.653) (20,0.658) (50,0.674) (100,0.682)};
\addplot+[mark=triangle*, thick] coordinates {(20,0.674) (50,0.691) (100,0.697) (200,0.717)};
\addplot+[mark=diamond*, dashed, thick] coordinates {(10,0.626) (20,0.624) (50,0.642) (100,0.637) (200,0.659)};
\legend{zero-shot, trained, trained-long, bge-reranker-v2-m3}
\end{axis}
\end{tikzpicture}
\caption{Quality vs.\ candidate count (data of Table~\ref{tab:scale}). Trained adapters keep improving as more candidates share one prefill ($k{=}200$ on the chunked path); the zero-shot adapter degrades beyond $k{=}20$. Peak memory grows from 15\,GiB ($k{=}10$) to 56\,GiB ($k{=}100$) because of per-branch cache copies.}
\label{fig:scale}
\end{figure}

Memory grows faster than linearly in $k$ (56\,GiB at $k{=}100$), because in the current implementation every probe branch continues from its own copy of the shared state's attention cache, so memory scales with $k\times$ context length. Under a 64\,GiB cap, the zero-shot single call runs out of memory at $k{=}200$ (43k tokens); Kev's chunked serving path (state once, branch rows in chunks) still runs $k{=}200$ and $k{=}300$ (65k tokens, 23\,GiB, about 22\,s per query), but the zero-shot adapter's quality collapses there (0.505 on 38 queries at $k{=}200$). \textbf{Batching across queries} does not help with the current implementation: packing $B$ queries ($k{=}20$) into one call lowers throughput from 2.87 queries/s ($B{=}1$) to 1.57 ($B{=}16$) and runs out of memory at $B{=}32$, with scores within bf16 noise of unbatched ones (mean $|\Delta p|$ 0.002). States of different lengths are left-padded, which forces an explicit mask instead of the flash kernel, and each branch row copies its state's cache, so batching multiplies the copies. Letting all branches attend to one shared cache with a block mask would remove this copy and make memory linear in context length; with it, a reranker over a 128k-token shared prefill becomes feasible, which we leave to future work.

\subsection{End to end: the same weights rerank and answer}\label{sec:e4}
On HotpotQA distractor~\cite{yang2018hotpotqa} (first 300 validation questions; 10 paragraphs per question, 2 gold; paragraphs trimmed to 220 words), one bf16 copy of the 9B backbone serves both roles: \sys{} reranks with the decision adapter on, then the same backbone answers greedily (at most 48 new tokens). We compare answer EM/F1 (official HotpotQA scoring) given (a) all 10 paragraphs, (b) top-3 by bge-reranker-v2-m3, (c) top-3 by \sys{}, answering with the adapter off, (b$'$)/(c$'$) the top-3 of (b)/(c), answering with the adapter left on, and (d) the 2 gold paragraphs.

\begin{table}[t]
\centering
\setlength{\tabcolsep}{4pt}
\begin{tabular}{@{}lccccc@{}}
\toprule
Context given to the generator & EM & F1 & R@2 & R@3 & rerank ms\\
\midrule
(a) all 10 paragraphs & 0.533 & 0.716 & -- & -- & --\\
(b) bge-reranker-v2-m3 top-3 & 0.500 & 0.683 & 0.863 & 0.918 & 29\\
(c) \sys{} top-3, adapter off & 0.543 & 0.718 & 0.762 & 0.865 & 378\\
\midrule
(b$'$) bge-reranker-v2-m3 top-3, adapter on & 0.547 & 0.702 & 0.863 & 0.918 & 29\\
(c$'$) \sys{} top-3, adapter on & \textbf{0.590} & \textbf{0.757} & 0.762 & 0.865 & 378\\
\midrule
(d) oracle: 2 gold paragraphs & 0.607 & 0.789 & -- & -- & --\\
\midrule
\multicolumn{6}{l}{\emph{trained adapter (separate B200 run)}}\\
(b$_\mathrm{t}$) bge-reranker-v2-m3 top-3, adapter off & 0.503 & 0.684 & 0.863 & 0.918 & 24\\
(c$_\mathrm{t}$) \sys{} trained top-3, adapter off & 0.553 & \textbf{0.729} & \textbf{0.897} & \textbf{0.962} & 274\\
(b$'_\mathrm{t}$) bge-reranker-v2-m3 top-3, trained adapter on & 0.510 & 0.682 & 0.863 & 0.918 & 24\\
(c$'_\mathrm{t}$) \sys{} trained top-3, trained adapter on & 0.577 & 0.747 & 0.897 & 0.962 & 274\\
\bottomrule
\end{tabular}
\caption{End-to-end RAG on HotpotQA distractor ($n{=}300$; packed\_all, $k{=}10$). Upper block: zero-shot adapter; lower block: trained adapter. Every row uses the same 9B generator; rows (b$'$) and (c$'$) answer with the adapter left on. R@$n$: recall of the gold paragraphs in the top $n$. Latencies: upper block on a shared RTX PRO 6000, lower block on a B200.}
\label{tab:e4}
\end{table}

Although the cross-encoder retrieves more gold paragraphs (R@2 0.863 vs.\ 0.762), the generator answers better from \sys{}'s top-3 (F1 0.718 vs.\ 0.683), consistent with the generator selecting context it can use; this adapter-off gap is not significant (below), and (c) is level with giving the generator all 10 paragraphs (0.716). Answering with the adapter left on raises F1 further to 0.757, within 0.032 of the gold-paragraph ceiling; its answers are shorter (2.5 vs.\ 3.1 words on average), which matches HotpotQA's short-answer format. To separate this answer-style effect from reranking, row (b$'$) answers from the cross-encoder's top-3 with the same adapter-on generator: \sys{}'s context still wins by $+0.055$ F1 (paired bootstrap 95\% CI $[+0.018,+0.093]$, 10{,}000 resamples) and $+0.043$ EM ($[+0.003,+0.083]$). With the adapter off the gap is $+0.035$ F1 ($[-0.003,+0.074]$), positive but not significant at $n{=}300$.

\paragraph{Real-time mode switching.} With the adapter unmerged, switching modes changes no weights. Over 20 alternations of a real 10-passage rerank (adapter on) and a real 24-token generation (adapter off), entering adapter-off took 11\,ms (median) and leaving it 2.8\,ms; every on-visit produced bit-identical probabilities and every off-visit the identical greedy answer, and allocated GPU memory stayed at 16.77\,GiB (14.78\,GiB backbone, 0.08\,GiB LoRA, 1.89\,GiB LM head).

\paragraph{Does the model still chat with the adapter on?} On general prompts (an explanation, a translation, an arithmetic word problem) adapter-on and adapter-off outputs were identical or nearly so, and the per-token NLL on 1{,}024 tokens of Wikipedia text rose only from 1.556 to 1.564 (perplexity 4.74 to 4.78). On these prompts the decision LoRA changes general generation little, though on HotpotQA it measurably shortens answers; whether a single adapted model can replace toggling needs broader evaluation.

\paragraph{Trained reranker, end to end.} With the trained adapter (lower block of Table~\ref{tab:e4}), \sys{} also retrieves more gold paragraphs than the cross-encoder (R@2 0.897 vs.\ 0.863), and under the same base generator (adapter off) its top-3 improves answer F1 from 0.684 to 0.729 ($+0.045$, paired bootstrap 95\% CI $[+0.014,+0.076]$). This comparison isolates the effect of selecting different evidence from any effect of the adapter on answer style. With the trained adapter left on for both contexts the gap is $+0.065$ ($[+0.038,+0.094]$), but the trained adapter-on F1 (0.747) is below the zero-shot one (0.757), suggesting that the two adapters affect answer style differently. The lower-recall-but-better-answers pattern of the zero-shot adapter is consistent with useful evidence selection, but it does not show that the model has learned answer utility; our training labels are passage relevance, not answer quality. The adapter-off cross-encoder and all-paragraph rows of this run differ from the upper block by at most 0.003 F1 presumably because of bf16 kernel differences between GPUs.

\subsection{Three modes on one backbone}\label{sec:multi}
Because both the general decision adapter and the reranking adapter are LoRA modules over the same frozen backbone, a single resident copy can serve three modes, chosen per request: \textbf{base} chat (all adapters off), \textbf{general} decisions (the general adapter and its pointer head), and \textbf{rerank} (the MS MARCO adapter and its own head). We load the backbone once, attach the second adapter as a named PEFT adapter with its own head, and cycle base\,$\to$\,general\,$\to$\,rerank for 10 rounds (Table~\ref{tab:multi}).

\begin{table}[t]
\centering
\begin{tabular}{@{}lll@{}}
\toprule
Mode & Check against the model loaded alone & Switch into mode (median)\\
\midrule
Base & greedy chat text identical on all visits (3 prompts $\times$ 10) & 4.2--6.4\,ms\\
General & probabilities bit-identical ($\max|\Delta p| = 0$) & 9.4\,ms\\
Rerank & probabilities bit-identical ($\max|\Delta p| = 0$) & 8.8\,ms\\
\bottomrule
\end{tabular}
\caption{One 9B backbone, three modes, switched per request. Memory: 16.85\,GiB resident (backbone 14.78, each LoRA 0.08, both heads 0.016, LM head 1.89\,GiB), unchanged after the loop (16.86\,GiB).}
\label{tab:multi}
\end{table}

Each extra decision skill costs about 0.1\,GiB. On the same inputs the two adapters' probabilities differ by up to 0.81, so the modes are genuinely distinct. \textbf{Skill retention.} On the 231 public items of JevBench~\cite{jevbench2026}, a benchmark of typed decision requests in easy, original and hard tiers (the hard tier holds adversarial and multi-step items), the general adapter scores 184/231 (79.7\%) and the reranking adapter, warm-started from it, 175/231 (75.8\%), a net loss of 9 items, 7 of them in the hard tier. Specialising one adapter for reranking therefore causes moderate forgetting of general decisions, which the multi-adapter setup sidesteps: each request uses the adapter trained for it.

\subsection{Activated adapter: one read for reranking and answering}\label{sec:alora}
In the runs above the LoRA is active on every token, so the state is read by an adapted model and its cache is not the base model's. For \sys{}-Activated, following Activated LoRA~\cite{greenewald2025alora}, we switch the adapter off for the state pass and on only for the probe branches (the invocation point is the first probe token), in both training and inference, by patching Kev's shared-prefix step. On a 0.8B Qwen3.5 check, the patched adapter-on state cache equals the base model's bitwise (48 tensors: attention KV, DeltaNet convolution and recurrent states; max $|\Delta|=0$, vs.\ 8.18 unpatched), while branch outputs still differ from the base.

\textbf{Quality.} We train the activated adapter for 150 steps (1{,}200 MS MARCO records, H100, 17 minutes) and a matched control with the LoRA on all tokens (same data, steps and seed). The activated adapter reaches 0.740 / 0.351 nDCG@10 on SciFact / NFCorpus (packed\_all, $k{=}20$), against 0.744 / 0.351 for the control: reranking scores are similar in this matched experiment.

\textbf{Chat.} Without probe tokens the activated model runs the exact base weights, so its greedy chat output is identical to the base by construction; the control adapter already changes the wording on one of two test prompts.

\textbf{Cache reuse.} Because the state cache is now the base model's own, generation can continue from it. On 100 HotpotQA distractor questions (10 paragraphs, median state 1.4k tokens, H100 not shared): the state pass takes 64\,ms, the 10 probes fork from its cache in 85\,ms (R@3 0.97), and the base model then answers by continuing the same cache with the question in the chat format. Continuing the cache gives EM/F1 0.61 / 0.750 against 0.59 / 0.754 for re-prefilling the top-3 passages, with a lower answer latency (138\,ms vs.\ 181\,ms median). On 5 checked questions the continued answer equals that of a fresh base-model prefill of the same text. Telling the model which passages scored highest added nothing at this scale, and a continuation without the chat turn markers is much worse (F1 0.577). In this mode the generator sees all passages; the reranking scores do not select its context. The passages are thus prefilled once for both reranking and answering; the continued cache still contains all 10 candidates. This is a 100-question, single-seed result.

\subsection{Order sensitivity}\label{sec:order}
In packed\_all every probe sees all passages, so scores can depend on passage order. On 100 SciFact queries (220-word passages, as in Table~\ref{tab:e3}) with the MS MARCO-trained adapter, we score each query's 20 candidates in BM25 order and in 4 random orders. nDCG@10 ranges from 0.678 (BM25 order) to 0.693 across orders (per-query standard deviation 0.020); averaging scores over the 5 orders gives 0.689. The median per-passage probability changes by $10^{-4}$ across orders (90th percentile 0.033, maximum 0.36), and the top-ranked passage is the same as under BM25 order in 91\% of cases. Order effects are therefore small on average but not negligible for individual passages, and the slightly lower score in BM25 order may indicate a mild position effect.

\subsection{Serving chat and decisions in one vLLM engine}\label{sec:vllm}
vLLM~\cite{kwon2023vllm} has no pointer head, but it can expose hidden states. We serve the base model with the reranking LoRA registered as a vLLM adapter: chat requests use no adapter, decision rows are one-token requests with the adapter, and the residual stream at the readout positions is passed through the final norm and the 256-dimensional pointer head outside the engine. Rows of one request share the state through vLLM's prefix cache on the hybrid model. In a single \texttt{generate} call, 4 chat answers (48 tokens each) and the 19 decision rows of one packed request completed in 0.76\,s, against 0.43\,s for the 4 chats alone. On 20 SciFact packed requests (400 probes), probabilities differ from Kev by 0.006 on average (maximum 0.10, about 2.6$\times$ the 0.039 spread between Kev's two bf16 inference paths, the default and the shared-prefix fast path), per-query Spearman rank correlation 0.995, with 4 of 400 decisions flipped; serving throughput on a shared B200 was 1.54 requests/s vs.\ 0.86 for Kev's sequential fast path. Production use needs a connector that exports only the readout positions and the head inside the engine.

\section{Limitations}
This is a proof of concept. Evaluation sets are small (hundreds of queries) and cover two BEIR datasets and one QA dataset. We test one backbone family (hybrid 9B). Latencies are from an unoptimised eager path; a dedicated cross-encoder remains faster per query. The shared-state construction itself is provided by the Kev runtime; our contribution is its use and training for reranking inside RAG with the generator's own weights. In packed\_all every probe sees all passages, so a passage's score depends somewhat on its neighbours and on order (\S\ref{sec:order}); averaging over orders is a simple remedy. BEIR differences are not significance-tested, and the per-passage baseline was not re-run with the trained adapter. Latencies come from different, partly shared GPUs (RTX PRO 6000, B200, H100), and speed-up ratios depend on load. The zero-shot adapter degrades beyond $k{=}20$ and collapses at $k{=}200$; batching across queries currently lowers throughput. The activated-adapter and cache-reuse results are single-seed, on 100 questions, the bitwise cache check used a 0.8B model, and with cache reuse the answer sees all passages rather than the reranked subset. Chat quality with the adapter on was checked on a handful of prompts, and specialising the adapter for reranking loses 9 JevBench items. Memory currently grows with $k\times$ context length (\S\ref{sec:scale}), and the vLLM path is a prototype whose probabilities differ from Kev's by 0.006 on average (maximum 0.10; 4 of 400 decisions flipped; \S\ref{sec:vllm}).

\section{Conclusion}
\sys{} shows that the generator can also serve as its reranker. Passage-specific probes share one candidate prefill, while lightweight adapters support ranking and general decisions on a single resident backbone. In the evaluated settings, the resulting rankings are competitive with a dedicated cross-encoder, and they improve downstream QA under the same generator. Probe-only adaptation further allows generation to reuse the original passage cache. The next questions are how these benefits extend across larger benchmarks and backbone families, how far a shared-cache implementation can push the candidate count toward 128k-token contexts, and how optimized serving systems change the latency trade-off against small cross-encoders.

\section*{Availability}
\begin{sloppypar}All adapters and code are released under Apache-2.0 by Airun Intelligence Labs (the repositories keep the working name GlanceRAG): the reranker (trained-long in the repository root, the 39-minute checkpoint under \texttt{checkpoints/msmarco-short}) at \url{https://huggingface.co/airun-ai/GlanceRAG-MiMo-9B-Reranker}, the activated variant at \url{https://huggingface.co/airun-ai/GlanceRAG-MiMo-9B-Reranker-Activated}, and the general decision adapter at \url{https://huggingface.co/airun-ai/MiMo-V2.6-Distill-Qwen-9B-Decision}. Code is in the \texttt{code/} folders of the two reranker repositories.\end{sloppypar}


\begin{thebibliography}{99}\small
\bibitem{an2025hyperrag} Y.~An, Y.~Cheng, S.~J.~Park, J.~Jiang. HyperRAG: Enhancing quality-efficiency tradeoffs in retrieval-augmented generation with reranker KV-cache reuse. arXiv:2504.02921, 2025.
\bibitem{asai2023selfrag} A.~Asai, Z.~Wu, Y.~Wang, A.~Sil, H.~Hajishirzi. Self-RAG: Learning to retrieve, generate, and critique through self-reflection. ICLR, 2024.
\bibitem{bajaj2016msmarco} P.~Bajaj et al. MS MARCO: A human generated machine reading comprehension dataset. arXiv:1611.09268, 2016.
\bibitem{jevbench2026} Benchmark Heaven. JevBench: a benchmark for decision models. Software, \url{https://github.com/fstandhartinger/jevbench}, commit 7625ecb1, 2026.
\bibitem{boteva2016nfcorpus} V.~Boteva, D.~Gholipour, A.~Sokolov, S.~Riezler. A full-text learning to rank dataset for medical information retrieval. ECIR, 2016.
\bibitem{chen2024bgem3} J.~Chen, S.~Xiao, P.~Zhang, K.~Luo, D.~Lian, Z.~Liu. BGE M3-Embedding: Multi-lingual, multi-functionality, multi-granularity text embeddings through self-knowledge distillation. Findings of ACL, 2024.
\bibitem{chen2024icr} S.~Chen, B.~J.~Guti\'errez, Y.~Su. Attention in large language models yields efficient zero-shot re-rankers. ICLR, 2025.
\bibitem{greenewald2025alora} K.~Greenewald et al. Activated LoRA: Fine-tuned LLMs for intrinsics. arXiv:2504.12397, 2025.
\bibitem{hu2022lora} E.~J.~Hu, Y.~Shen, P.~Wallis, Z.~Allen-Zhu, Y.~Li, S.~Wang, L.~Wang, W.~Chen. LoRA: Low-rank adaptation of large language models. ICLR, 2022.
\bibitem{kwon2023vllm} W.~Kwon, Z.~Li, S.~Zhuang, Y.~Sheng, L.~Zheng, C.~H.~Yu, J.~E.~Gonzalez, H.~Zhang, I.~Stoica. Efficient memory management for large language model serving with PagedAttention. SOSP, 2023.
\bibitem{liu2024lost} N.~F.~Liu, K.~Lin, J.~Hewitt, A.~Paranjape, M.~Bevilacqua, F.~Petroni, P.~Liang. Lost in the middle: How language models use long contexts. TACL, 2024.
\bibitem{lu2024turborag} S.~Lu, H.~Wang, Y.~Rong, Z.~Chen, Y.~Tang. TurboRAG: Accelerating retrieval-augmented generation with precomputed KV caches for chunked text. arXiv:2410.07590, 2024.
\bibitem{nogueira2020monot5} R.~Nogueira, Z.~Jiang, R.~Pradeep, J.~Lin. Document ranking with a pretrained sequence-to-sequence model. Findings of EMNLP, 2020.
\bibitem{kev2026} J.~Palmer. Kev: typed decision models. Software, \url{https://github.com/jaredpalmer/kev}, commit 0a6d3faf, 2026.
\bibitem{pradeep2023rankzephyr} R.~Pradeep, S.~Sharifymoghaddam, J.~Lin. RankZephyr: Effective and robust zero-shot listwise reranking is a breeze! arXiv:2312.02724, 2023.
\bibitem{ratner2023pcw} N.~Ratner, Y.~Levine, Y.~Belinkov, O.~Ram, I.~Magar, O.~Abend, E.~Karpas, A.~Shashua, K.~Leyton-Brown, Y.~Shoham. Parallel context windows for large language models. ACL, 2023.
\bibitem{reddy2024first} R.~G.~Reddy, J.~Doo, Y.~Xu, M.~A.~Sultan, D.~Swain, A.~Sil, H.~Ji. FIRST: Faster improved listwise reranking with single token decoding. EMNLP, 2024.
\bibitem{shi2023replug} W.~Shi, S.~Min, M.~Yasunaga, M.~Seo, R.~James, M.~Lewis, L.~Zettlemoyer, W.~Yih. REPLUG: Retrieval-augmented black-box language models. NAACL, 2024.
\bibitem{sun2023rankgpt} W.~Sun, L.~Yan, X.~Ma, S.~Wang, P.~Ren, Z.~Chen, D.~Yin, Z.~Ren. Is ChatGPT good at search? Investigating large language models as re-ranking agents. EMNLP, 2023.
\bibitem{thakur2021beir} N.~Thakur, N.~Reimers, A.~R\"uckl\'e, A.~Srivastava, I.~Gurevych. BEIR: A heterogeneous benchmark for zero-shot evaluation of information retrieval models. NeurIPS Datasets and Benchmarks, 2021.
\bibitem{wadden2020scifact} D.~Wadden, S.~Lin, K.~Lo, L.~L.~Wang, M.~van Zuylen, A.~Cohan, H.~Hajishirzi. Fact or fiction: Verifying scientific claims. EMNLP, 2020.
\bibitem{yang2018hotpotqa} Z.~Yang, P.~Qi, S.~Zhang, Y.~Bengio, W.~W.~Cohen, R.~Salakhutdinov, C.~D.~Manning. HotpotQA: A dataset for diverse, explainable multi-hop question answering. EMNLP, 2018.
\bibitem{yang2025gdn} S.~Yang, J.~Kautz, A.~Hatamizadeh. Gated delta networks: Improving Mamba2 with delta rule. ICLR, 2025.
\bibitem{yao2024cacheblend} J.~Yao, H.~Li, Y.~Liu, S.~Ray, Y.~Cheng, Q.~Zhang, K.~Du, S.~Lu, J.~Jiang. CacheBlend: Fast large language model serving for RAG with cached knowledge fusion. EuroSys, 2025.
\bibitem{zhang2025qwen3emb} Y.~Zhang et al. Qwen3 Embedding: Advancing text embedding and reranking through foundation models. arXiv:2506.05176, 2025.
\end{thebibliography}
\end{document}